\documentclass[11pt]{article}

\usepackage[final]{acl}

\usepackage{times}
\usepackage{latexsym}

\usepackage[T1]{fontenc}

\usepackage[utf8]{inputenc}

\usepackage{microtype}

\usepackage{inconsolata}

\usepackage{graphicx}
\usepackage{booktabs}
\usepackage{multirow} 
\usepackage{amsmath}
\usepackage{tikz}
\usetikzlibrary{arrows.meta,fit,positioning}
\title{Efficient Agentic Audio-Visual Question Answering through Toolset Pruning and Modality Consolidation}

\author{
  Yi Wang$^{1}$ \quad
  Tianzi Wang$^{2}$ \quad
  Peter Bell$^{1}$ \\[2pt]
  $^{1}$School of Informatics, University of Edinburgh, UK \\
  $^{2}$Department of Systems Engineering and Engineering Management, \\
  Chinese University of Hong Kong, Hong Kong SAR \\[2pt]
  \texttt{\{Wang.Yi, Peter.Bell\}@ed.ac.uk} \quad
  \texttt{twang@se.cuhk.edu.hk}
}

\begin{document}
\maketitle
\begin{abstract}
Fine-grained audio-visual question answering (AVQA) requires models to resolve complex cross-modal correspondences and temporally grounding from sparsely distributed evidence. Omni-modal agents, which autonomously orchestrate perception tools to actively collect evidence, achieve strong performance, but it remains unclear whether active perception from both audio and visual modalities depends on broad specialized toolsets, especially audio or visual modality-specific tools. Such systems may also introduce long execution trajectories and substantial inference cost.
We investigate whether comparable performance can be retained with a compact workflow exposing only essential perceptual capabilities. Starting from the specialized-tool OmniAgent, we conduct staged controlled experiments that prune the toolset and progressively consolidate modality-specific tools into omni-modal tools, motivating a two-tool agent comprising only Omni-modal captioning and Omni-modal QA. Across three widely used AVQA benchmarks, our compact agent achieves comparable or better accuracy than the specialized-tool agent while reducing inference cost by around 70\%.
\end{abstract}

\section{Introduction}
Audio-visual question answering (AVQA) requires models to answer questions depending jointly on the auditory and visual streams of a video \cite{zhou2025daily, yang2025audio, hong2025worldsense}. Fine-grained AVQA is particularly challenging because relevant evidence may be temporally sparse and distributed across modalities, requiring models to identify temporally aligned events and resolve complex cross-modal correspondences \citep{chowdhury2024meerkat,sung2024avhbench}. Recent omni-modal large language models (OmniLLMs), which jointly process audio, video, and text, have achieved strong performance on AVQA \citep{tang2025videosalmonn2,xu2025qwen3omni,yang2025humanomniv2, comanici2025gemini,team2026qwen3}.

Nevertheless, running single-pass inference with OmniLLMs may miss sparse, question-relevant evidence distributed across a video \citep{xu2026agentic}, which becomes more pronounced in long-video understanding \citep{zhou2025mlvu,fu2025videomme}. This limitation has motivated agentic systems that actively acquire evidence through iterative planning, temporal localization, and external perception tools \cite{wang2024videoagent,zhang2026deep,wang2025active, tao2025omniagent,xing2026native, xu2026agentic}.
Recent audio-visual agents have shown that such active perception can outperform direct inference with the corresponding OmniLLMs \cite{tao2025omniagent, xing2026native}.
Meanwhile, these systems often expose the planner to a broad collection of function-specific tools \cite{tao2025omniagent, li2026videothinker, fan2026tool}. For example, OmniAgent \cite{tao2025omniagent} adopts a global audio captioning tool, which generates a caption from the audio track of the entire video, along with tools for speech recognition, event localization, global video understanding, and clip-level video understanding.
Flexible orchestration over such toolsets may improve coverage of retrieving specific evidence, but may also introduce long execution trajectory, redundant reasoning and observations, and substantial inference cost.

This motivates us to investigate how much of the performance depends on maintaining a broad set of specialized perception tools, especially in the context that OmniLLMs capacity kept evolving.
In particular, we raise two specific research questions. 1) Tool selection: can we prune the broad specialized toolset to substantially fewer tools without sacrificing accuracy. 2) Modality merging: after identifying the tools responsible for necessary capabilities, can we consolidate modality-specific audio tools and visual tools into omni-modal tools to improve efficiency? 

We investigate these questions through a staged study of tool type selection and modality consolidation. We take OmniAgent \cite{tao2025omniagent} as an initial instance, which achieved SOTA performance on AVQA among agentic methods, first ablating its broad toolsets consisting of function-specific tools to identify which are necessary. We then progressively replace modality-specific captioning and question-answering tools with omni-modal tools with corresponding roles under controlled experiments. 
This process leads to a compact agentic system with two complementary tools: the Omni Captioning Tool, which obtains broad context, and the Omni QA Tool, which conducts question-directed evidence acquisition.

Our compact two-tool agent consistently outperforms its corresponding direct OmniLLM baseline, Gemini-2.5-Flash, and achieves strong performance across Daily-Omni, AVUT, and WorldSense (78.86\%, 80.57\%, and 58.39\%). It also achieves comparable or better accuracy than the full specialized-tool method OmniAgent, while reducing tool calls by 72--79\% and lowering estimated inference cost by 72--75\%.

Our controlled ablations further show that, first, a broad collection of modality-specific tools is not necessary for strong agentic AVQA: global audio-visual captioning and question-directed perception capture most of the required capabilities. Second, consolidating these capabilities into omni-modal tools yields a compact system with fewer tool calls. Third, the agentic audio-visual understanding pipeline remains beneficial with a strong OmniLLM as the shared perception backbone, with the benefit being more prominent in long videos.

Our main contributions include the followings. 
\begin{itemize}
    \item  We propose a compact two-tool agentic audio-visual understanding system that preserves strong performance while substantially reducing tool calls, context consumption, and estimated inference cost.
    \item We conduct controlled capability-removal and modality-consolidation studies, showing that strong and efficient agentic AVQA depends more on retaining a small set of complementary perceptual capabilities than on providing an extensive specialized toolset.
\end{itemize}

\section{Related Work}
\begin{figure*}[t]
    \centering
    \includegraphics[width=0.8\textwidth]{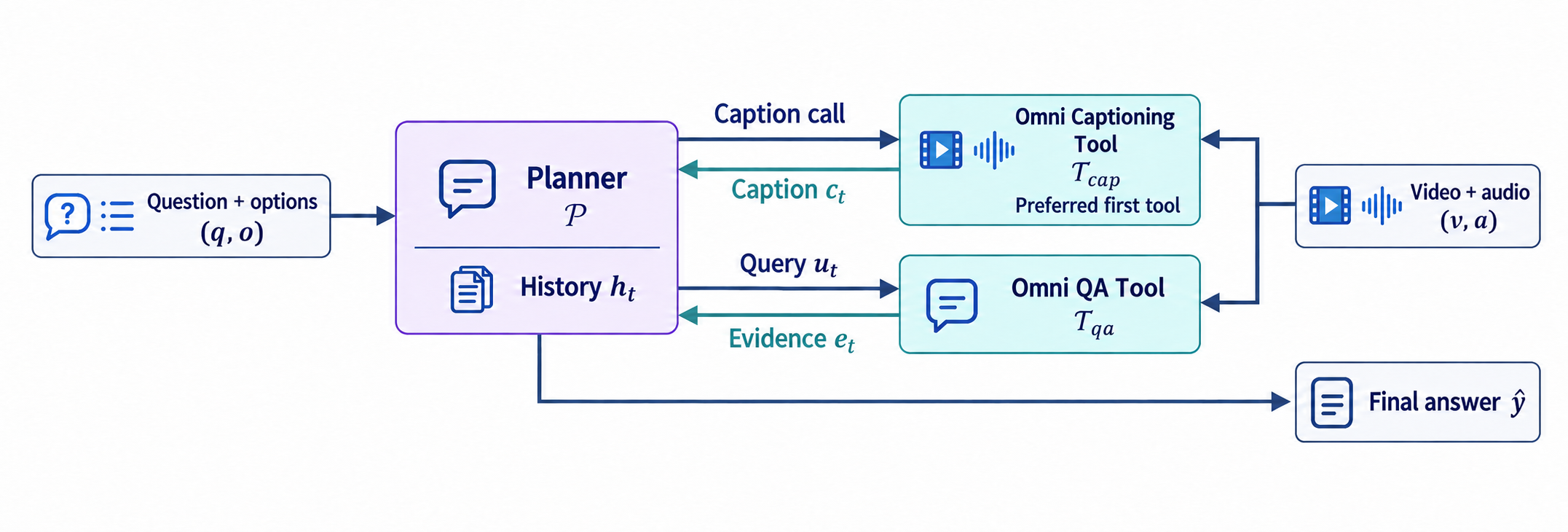}
    \vspace{-0.5em}
    \caption{\textbf{Overview of the proposed framework.} During inference, a text-only planner dynamically selects between two omni-modal tool interfaces or returns the final answer. A prompt-level caption-first policy provides initial global context without fixing the remainder of the evidence-acquisition trajectory.}
    \label{fig:pipeline}
\vspace{-1em}
\end{figure*}

\subsection{Audio-Visual Question Answering}
\label{ssec:avqa}
In audio-visual question answering (AVQA), a model must answer questions that depend jointly on the auditory and visual streams of a video \citep{zhou2025daily,yang2025audio,hong2025worldsense}. 
Due to the key role of both the audio and the video modality information in AVQA, it evaluates not only modality-specific perception, but also the ability to locate sparsely distributed evidence, establish temporal alignment, and reason across the retrieved audio-visual information.

Earlier modular approaches commonly externalize video content into text by aggregating segment-level captions, followed by combining textual evidence with a separate language reasoner \citep{hu2023promptcap,du2023zero,gou2025rapid}. Among these works modularized video understanding into separate captioning and reasoning modules, some developed their methods with an emphasis on generating question-conditioned descriptions, which already demonstrated the importance of question adaptive evidence acquisition \citep{hu2023promptcap}.
More recently, OmniLLMs have unified audio, video, and text processing within a single model, enabling direct end-to-end AVQA \citep{xu2025qwen3omni,yang2025humanomniv2,comanici2025gemini,team2026qwen3}. Beyond serving as strong direct-inference baselines, these models are also adopted by agentic systems as perception tools that acquire question-relevant audio-visual evidence iteratively \cite{tao2025omniagent, xing2026native}.

\subsection{Agentic AVQA}
Prior agentic video-understanding works have used question-guided frame selection, hierarchical video clipping, and question decomposition to make relevant evidence easier to find and reason over \citep{wang2024videoagent,wang2024videotree,zhang2024omagent,liao2025divide}. Another noticeable line of agentic video-understanding systems adopts ReAct-style active-perception loops, in which a planner, typically a text LLM, repeatedly determines what evidence is missing, invokes an appropriate video retrieval or perception tool, and updates its reasoning from the returned observations \citep{zhang2026deep,wang2025active}.  These studies demonstrate the value of agentic video-understanding systems, particularly in terms of actively acquiring evidence according to the question and task decomposition. 

Audio-visual understanding additionally requires agents to determine which modality to inspect and how evidence from audio and video should be temporally aligned. One approach keeps evidence-acquisition operations separated by modality.
The OmniAgent system of \citet{tao2025omniagent} (hereafter, \emph{specialized-tool OmniAgent}) orchestrates separate tools for audio captioning, QA, ASR, event localization, and video QA, using audio evidence to guide subsequent visual inspection. OmniRAG-Agent likewise retrieves evidence from separate image and ASR-indexed audio banks \citep{zhu2026omnirag}, while emphasizing more on incorporating RAG mechanism into the system.
Other approaches rely more directly on omni-modal perception. AOP-Agent uses an OmniLLM to build a hierarchical audio-visual memory, and focus on workflow design where an agent explores through multiple observation operations \citep{xu2026agentic},  
whereas the OmniAgent system of \citet{xing2026native} (hereafter, \emph{native OmniAgent}) trains a single OmniLLM to perceive requested frames, audio segments, and synchronized clips.
These systems differ not only in how they organize perceptual tools, but also in their perception model backbones, training, and workflows. Consequently, their reported results leave the question open whether agentic AVQA benefits more from a broad set of modality-specific functions or from a compact set of essential OmniLLM-based tools under a matched workflow.

\section{Method}
\label{sec:method}

\subsection{Compact Agentic Audio-Visual Reasoning}
\label{ssec:agent-workflow}

Let $v$ and $a$ denote the visual and audio streams of a video, and let
$q$ and $o$ denote a question and its answer options. Our system uses a
text-only LLM planner (hereafter, the planner), denoted by
$\mathcal P$. The planner does not directly access the raw audio or video.
Instead, it processes the question, answer options, and textual tool
observations; constructs targeted QA queries; selects subsequent actions;
and returns the final answer.

The planner acquires evidence through two tools. The Omni Captioning Tool
$\mathcal T_{\mathrm{cap}}$ provides broad audio-visual context, including
salient events, with each event description sentence accompanied by its time stamp. The Omni QA Tool
$\mathcal T_{\mathrm{qa}}$ answers a targeted QA query about the audio-visual
content. Their observations are
\(c_t = \mathcal T_{\mathrm{cap}}(v,a;p_{\mathrm{cap}})\),
and 
\(e_t = \mathcal T_{\mathrm{qa}}(v,a,u_t;p_{\mathrm{qa}})\),
where $u_t$ is a QA query raised by planner, $c_t$ is a caption observation,
and $e_t$ is a targeted evidence observation. 

At each step, the planner selects an action
\(z_t = \mathcal P(q,o,h_t;p_{\mathrm{plan}})\), with $z_t$ is either calling captioning tool, QA tool or making decison, $h_t$ being the textual interaction history and $p_{\mathrm{plan}}$ is the
planner instruction. The quantities $p_{\mathrm{cap}}$, $p_{\mathrm{qa}}$, and
$p_{\mathrm{plan}}$ are natural-language instructions passed to the corresponding models. The planner instruction impose a soft constraint that the Omni Captioning Tool should be called first. After the initial caption, the planner dynamically calls either tool again or return $\hat y$ when obtaining adequate evidence. Tool queries and observations are appended to $h_t$ before the next decision. If the valid-call budget is exhausted, the planner must answer from the evidence already collected.

\subsection{Capability-Removal Study}
\label{ssec:capability-ablation}
The workflow described in \ref{ssec:agent-workflow} is proposed based on a series of controlled experiments on toolset pruning.
We first study which capabilities of a broad specialized toolkit
should be retained. We group the tools in specialized-tool
OmniAgent into five functional categories: global audio captioning
($C$); question-directed audio and video answering ($Q$, comprising
\texttt{audio\_qa}, \texttt{video\_global\_qa}, and
\texttt{video\_clip\_qa}); audio-event listing and localization
($E$); speech recognition ($A$); and video metadata ($M$).
Starting from the full $C+Q+E+A+M$ toolkit, we remove each group
in turn on the same 125-question validation subset. We also examine
the combined removal of $A$ and $E$, and a minimal $C+Q$
configuration. Grouping tools by function allows us to test whether
an apparent need for many individual tools reflects distinct
perceptual capabilities or redundant ways of acquiring evidence.

\subsection{Workflow and Modality-Consolidation Study}
\label{ssec:modality-ablation}

The second study begins with the retained specialized
$C+Q+M$ configuration and separates workflow changes from tool
changes. We first reproduce this configuration in a common
execution framework while preserving its tool roles and
OmniAgent-derived instructions as closely as possible. We then
remove the instruction encouraging cross-tool verification before
replacing the planner instructions with a simpler evidence-driven
policy. Both policies remain iterative: after each observation, the
planner can call another tool or return a final answer.

We next replace global audio captioning with the Omni Captioning Tool (also on the whole video),
initially leaving the order of tool calls unconstrained. Adding a planner
instruction to invoke the captioning tool later in a separate configuration isolates the effect of obtaining broad audio-visual context before question-directed
inspection. We then replace the three modality-specific QA tools with
omni-modal global and clip-level QA tools. Finally, we examine whether
planner-visible metadata and clip-level QA remain necessary, resulting in the
two-tool configuration containing only the Omni Captioning Tool and the Omni
QA Tool. 

All configurations use the same evaluation subset and retain the planner, history handling, and final-answer procedure across matched comparisons. The staged design of ablation study distinguishes changes in workflow, caption modality, QA-tool specialization, and tool availability.

\section{Experiments}
\label{sec:experiments}

We evaluate the final compact agent on Daily-Omni, AVUT, and WorldSense. We use
Daily-Omni for controlled tool and workflow ablations, and AVUT and WorldSense
to evaluate the resulting design across different content and duration
distributions. We compare against both direct OmniLLM inference and a
replication of the specialized-tool OmniAgent system.

\subsection{Evaluation Datasets}
\label{ssec:exp_data}

\paragraph{Daily-Omni.}
Daily-Omni~\cite{zhou2025daily} is our primary benchmark for workflow analysis.
It contains 1,197 audio-visual multiple-choice questions over videos no longer
than 60 seconds. We report final performance on the full benchmark. For the
staged ablations, we use a fixed 125-example validation subset stratified by task type to preserve approximately the distribution of the complete
dataset. The same subset is used for every ablation configuration. Detailed split statistics are provided in Appendix~\ref{app:daily_omni_statistics}.

\paragraph{AVUT and WorldSense.}
We evaluate all 1,734 questions in the human-annotated AVUT QA partition
\citep{yang2025audio} and the 3,172-question WorldSense evaluation set
\citep{hong2025worldsense}. Their duration distributions are reported in
Table~\ref{tab:duration_analysis}. The final configuration was fixed using
Daily-Omni before evaluation on AVUT and WorldSense. We included an additional WorldSense experiment in Appendix~\ref{app:worldsense_clip_qa_duration},
where the tool set includes a QA tool based on omni-modal video clip, which is a post-hoc generalization analysis and was not used to select the reported final system. Additional benchmark and artifact documentation is provided in Appendix~\ref{app:benchmark_documentation}.

\begin{table*}[t]
  \centering
  \small
  \setlength{\tabcolsep}{3pt}
  \renewcommand{\arraystretch}{1.04}
  \begin{tabular*}{\textwidth}{@{\extracolsep{\fill}}llrrrrrr@{}}
    \toprule
    Method & Inference paradigm & \# Tools & Acc. (\%) $\uparrow$
      & Calls $\downarrow$ & Input / output tok. $\downarrow$
      & Est. cost (\$) $\downarrow$ & Latency (s) $\downarrow$ \\
    \midrule
    \multicolumn{8}{l}{\textit{Daily-Omni} ($n=1{,}197$)} \\
    Gemini-2.5-Flash & Direct OmniLLM & -- & 77.03 & 0.000
      & 5.048k / 0.006k & 0.00154 & -- \\
    OmniAgent & Specialized agent & 8 & 77.53 & 7.270
      & 49.661k / 7.462k & 0.07142 & 118.32 \\
    Ours & Compact agent & \underline{2} & \textbf{78.86}
      & \underline{1.550} & \underline{13.000k / 1.670k}
      & \underline{0.01910} & \underline{14.90} \\
    \midrule
    \multicolumn{8}{l}{\textit{AVUT} ($n=1{,}734$)} \\
    Gemini-2.5-Flash & Direct OmniLLM & -- & 79.35 & 0.000
      & 11.457k / 0.149k & 0.00437 & -- \\
    OmniAgent & Specialized agent & 8 & 77.16 & 6.382
      & 33.898k / 4.272k & 0.08312 & 107.76 \\
    Ours & Compact agent & \underline{2} & \textbf{80.57}
      & \underline{1.777} & \underline{16.814k / 1.432k}
      & \underline{0.02047} & \underline{27.55} \\
    \midrule
    \multicolumn{8}{l}{\textit{WorldSense} ($n=3{,}172$)} \\
    Gemini-2.5-Flash & Direct OmniLLM & -- & 56.43 & 0.000
      & 22.517k / 0.303k & 0.00874 & -- \\
    OmniAgent & Specialized agent & 8 & \textbf{59.30} & 7.358
      & 69.986k / 5.558k & 0.11414 & 143.64 \\
    Ours & Compact agent & \underline{2} & 58.39
      & \underline{1.986} & \underline{39.017k / 2.019k}
      & \underline{0.03139} & \underline{31.29} \\
    \bottomrule
  \end{tabular*}
  \caption{Accuracy and inference efficiency across three audio-visual QA
  benchmarks. All direct Gemini-2.5-Flash and OmniAgent results are obtained
  from our own runs; OmniAgent is reproduced based on its released code.
  All efficiency values are averages per question. Bold indicates
  the highest accuracy within each benchmark; underlining marks the more
  efficient agentic system. Input/output tokens include all model calls.
  Latency is the mean per-question retry-adjusted wall-clock time. }
  \label{tab:main_cross_benchmark}
\end{table*}

\begin{table*}[t]
  \centering
  \small
  \setlength{\tabcolsep}{8pt}
  \renewcommand{\arraystretch}{1.04}
  \begin{tabular*}{\textwidth}{@{\extracolsep{\fill}}lrrrrr@{}}
    \toprule
    Method & Overall & $\leq 1$ min & 1--3 min & 3--5 min & $>5$ min \\
    \midrule
    \multicolumn{6}{l}{\textit{AVUT} (bucket $n$: 1,125 / 571 / 7 / 31)} \\
    Gemini-2.5-Flash & 79.35 & 78.22 & 81.79 & \textbf{85.71} & 74.19 \\
    OmniAgent (reproduced) & 77.16 & 75.82 & 79.86
      & \textbf{85.71} & 74.19 \\
    Ours & \textbf{80.57} & \textbf{79.11} & \textbf{83.19}
      & \textbf{85.71} & \textbf{83.87} \\
    $\Delta$ (ours $-$ direct) & $+$1.22 & $+$0.89 & $+$1.40
      & $+$0.00 & \textbf{$+$9.68} \\
    \midrule
    \multicolumn{6}{l}{\textit{WorldSense} (bucket $n$: 1,059 / 1,386 / 348 / 379)} \\
    Gemini-2.5-Flash & 56.43 & \textbf{57.70} & 58.80 & 53.45 & 46.97 \\
    OmniAgent (reproduced) & \textbf{59.30} & 57.13 & \textbf{60.82}
      & 54.02 & \textbf{64.64} \\
    Ours & 58.39 & 55.43 & 59.52 & \textbf{56.90} & 63.85 \\
    $\Delta$ (ours $-$ direct) & $+$1.96 & $-$2.27 & $+$0.72
      & $+$3.45 & \textbf{$+$16.88} \\
    \bottomrule
  \end{tabular*}
  \caption{Accuracy by video duration for direct Gemini-2.5-Flash inference, reproduced OmniAgent, and our compact two-tool agent. }
  \label{tab:duration_analysis}
\end{table*}

\subsection{Planner and Shared Perception Backbone}
\label{ssec:exp_models}

Our agent pipeline is implemented in DSPy~\citep{khattab2024dspy} and uses
o3\footnote{OpenAI o3 and o4-mini System Card:
\url{https://openai.com/index/o3-o4-mini-system-card/}} as the text-only planner. The final compact system uses medium reasoning and permits at
most six valid tool calls. It first obtains an omni-modal caption,
then either requests further evidence or returns the final answer.

Gemini-2.5-Flash serves as the frozen, shared perception backbone for both
tools. Such model selection is because specialized-tool OmniAgent system, which we took as starting point of our tool selection controlled experiment, reported its best performance when all tools were based on Gemini-2.5-Flash model. Despite the shared OmniLLM backbone, different tools have their roles specified just by different prompts.
The Omni Captioning Tool returns a broad, timestamped description,
whereas the Omni QA Tool answers a QA query with audio-visual inputs generated by
the planner. We run all direct Gemini-2.5-Flash baselines ourselves, sending
the raw audio-video input, question, and options to the model in a single call.
These runs use the same evaluation inputs, media preprocessing, failure
handling, and scoring protocol as the agent systems; only the inference
paradigm differs.
Complete inference parameters and the planner and tool prompts are provided in
Appendix~\ref{app:implementation_details}.

\vspace{-0.2cm}
\subsection{OmniAgent Replication and Ablation Protocol}
\label{ssec:ablation_protocol}
\vspace{-0.1cm}

We replicate specialized-tool OmniAgent from its released
implementation~\cite{tao2025omniagent}, retaining its dynamic ReAct-style
workflow and eight tools spanning global audio captioning, audio QA, speech recognition,
audio-event analysis, global and clip-level video QA, and video metadata. The
replication uses an o3 planner and Gemini-2.5-Flash-based perception components, because such model choice exerts the best performed system reported in \citet{tao2025omniagent}. The \textit{specialized-tool OmniAgent} uses high reasoning, whereas our final agent uses medium reasoning; we isolate this change in the ablation study.

All controlled ablations use the same stratified 125-example Daily-Omni subset.
Within each matched comparison throughout the ablation experiments, we keep the examples, model backbones, history handling, retry policy, and final-answer procedure fixed. For the tool pruning ablation experiments, each configuration is run three times and we report mean and sample standard
deviation, so as to mitigate the influence from run-to-run variance of closed-source model. We reported the full results on three benchmarks by a single run with fixed seed, as to save the experiment costs and align with the practice of previously reported AVQA results. 

\subsection{Evaluation and Efficiency Accounting}
\label{ssec:exp_metrics}

We report exact multiple-choice accuracy over the complete evaluation set. 
A valid tool call is an executor-accepted invocation that returns an observation; each invocation of either the Omni Captioning Tool or the Omni QA Tool counts once, including repeated invocations. Parser-rejected or malformed actions, and automatic retries to handle erroneous API calls are not counted into the valid call budget. The planner's final-answer round is excluded for the tool calling counts, but the tokens consumed in such round are taken into account. Input and output tokens sum the corresponding tokens from both the planner and the perception-model tokens. 
We average tokens, calls, and estimated cost per question and apply the same definitions to the OmniAgent traces.
Estimated cost applies the standard API price schedule at the time of the experiments to the recorded usage.

\begin{table*}[t]
  \centering
  \small
  \setlength{\tabcolsep}{8pt}
  \renewcommand{\arraystretch}{1.03}
  \begin{tabular}{llrrrr}
    \toprule
    Setting & Included tool groups & Acc. (\%) $\uparrow$
      & Calls $\downarrow$ & Input / output tok. $\downarrow$
      & Est. cost (\$) $\downarrow$ \\
    \midrule
    $-$A & C + Q + E + M & 77.60 $\pm$ 2.12
      & 7.667 & 31.133k / 4.284k & 0.05729 \\
    $-$E & C + Q + A + M & 78.67 $\pm$ 0.92
      & 7.853 & 30.891k / 4.911k & 0.06226 \\
    \textbf{$-$A$-$E} & \textbf{C + Q + M} & \textbf{78.93 $\pm$ 1.85}
      & 7.973 & 30.386k / 4.470k & 0.05867 \\
    $-$C & Q + E + A + M & 75.47 $\pm$ 1.67
      & 6.445 & 26.264k / 4.101k & 0.05258 \\
    $-$Q & C + E + A + M & 62.13 $\pm$ 2.57
      & 5.005 & 15.639k / 4.103k & 0.04573 \\
    $-$M & C + Q + E + A & 78.13 $\pm$ 2.31
      & 6.568 & 26.398k / 4.360k & 0.05488 \\
    Minimal & C + Q & 74.13 $\pm$ 2.01
      & 6.683 & 23.791k / 3.913k & 0.04936 \\
    \bottomrule
  \end{tabular}
  \caption{Stage 1 tool-group ablation on the selected Daily-Omni validation subset ($n=125$). C is global audio captioning; Q groups all QA tools; E groups audio-event listing and localization; A is automatic speech recognition; and M is video metadata. Accuracy and efficiency metrics are mean $\pm$ sample standard deviation over three runs. }
  \label{tab:stage1_tool_selection}
\end{table*}

\begin{table*}[t]
  \centering
  \small
  \setlength{\tabcolsep}{5pt}
  \renewcommand{\arraystretch}{1.05}
  \begin{tabular*}{\textwidth}{
    @{\extracolsep{\fill}}llrrrr@{}
  }
    \toprule
    System & Controlled change
      & Acc. (\%) $\uparrow$
      & Calls $\downarrow$
      & Input / output tok. $\downarrow$
      & Est. cost (\$) $\downarrow$ \\
    \midrule

    \multicolumn{6}{@{}l}{
      \textit{Stage 2: Workflow and modality consolidation}
    } \\
    \addlinespace[2pt]

    0 & Port to DSPy implementation
      & 75.47 $\pm$ 1.22
      & 4.309 & 19.783k / 6.414k & 0.08046 \\

    1 & Gemini deterministic sampling
      & 76.00 $\pm$ 1.60
      & 4.232 & 19.771k / 6.261k & 0.07797 \\

    2 & Adopt compact workflow
      & 72.27 $\pm$ 1.22
      & 3.005 & 11.344k / 3.551k & 0.04293 \\

    3 & Global audio captioning $\rightarrow$ Omni Captioning Tool
      & 75.73 $\pm$ 1.67
      & 2.547 & 15.520k / 4.458k & 0.05033 \\

    4 & Add caption-first instruction
      & 77.87 $\pm$ 0.46
      & 2.397 & 14.003k / 3.722k & 0.03988 \\

    5 & Specialized QA tools $\rightarrow$ omni-modal QA tools
      & 74.40 $\pm$ 1.39
      & 2.013 & 14.434k / 2.909k & 0.03223 \\

    6 & Hide video metadata
      & 78.40 $\pm$ 1.60
      & 1.941 & 13.931k / 2.684k & 0.02949 \\

    \midrule

    \multicolumn{6}{@{}l}{
      \textit{Stage 3: Further compacting the omni-tool workflow}
    } \\
    \addlinespace[2pt]

    7 & Reduce call budget
      & 77.60 $\pm$ 4.45
      & 1.813 & 15.274k / 2.712k & 0.02803 \\

    8 & Ablate clip QA
      & 77.60 $\pm$ 3.20
      & 1.627 & 13.939k / 2.549k & 0.02675 \\

    \textbf{9$^{*}$ (ours)} & Lower reasoning effort
      & 77.07 $\pm$ 2.44
      & \textbf{1.547}
      & \textbf{12.161k / 1.648k}
      & \textbf{0.01791} \\

    10$^{*}$ & Remove caption-first instruction from Sys. 9
      & 74.93 $\pm$ 1.67
      & 1.568 & 13.745k / 1.892k & 0.02183 \\

    11$^{*}$ & Add back clip-level Omni QA from Sys. 9
      & 76.80 $\pm$ 1.39
      & 1.768 & 16.428k / 1.976k & 0.02211 \\

    \bottomrule
  \end{tabular*}

  \caption{
    Combined Stage 2 and Stage 3 ablations on the selected Daily-Omni
    validation subset ($n=125$). Accuracy is reported as mean $\pm$ sample
    standard deviation over three runs; 
    $^{*}$ indicates medium planner reasoning effort; all other settings use
    high reasoning effort.
  }
  \label{tab:stage2_stage3_ablation}
\end{table*}

\section{Results}
\label{sec: results}

\subsection{Performance and Efficiency of the Compact Agent}
\label{ssec: overall_results}
Table~\ref{tab:main_cross_benchmark} shows that the efficiency gains of the compact workflow do not come at the expense of overall accuracy. 
Compared with the reproduced full OmniAgent, our two-tool agent improves accuracy by 1.33 and 3.41 percentage points on Daily-Omni and AVUT, 
respectively, while remaining within 0.91 points on WorldSense. Meanwhile, it reduces average tool calls by 72.2--78.7\%, input-token consumption by at least 44\%, output-token consumption by at least 63\%, and estimated inference cost by at least 71\%, which are substantial reductions. Compared with single-pass Gemini model, the agentic workflow consistently improves overall accuracy across all three benchmarks, though at the cost of more tool calls and tokens consumed.
The same accuracy trend holds on a video-disjoint Daily-Omni held-out set (Appendix~\ref{app:daily_omni_heldout} and Table~\ref{tab:daily_omni_video_disjoint}).
Our latency is also lower than that of the reproduced OmniAgent, consistent with the other efficiency metrics.

\subsection{Agent Performance by Video Durations}
To examine when this additional computation is most useful, Table~\ref{tab:duration_analysis} further breaks down performance by video duration, particularly for WorldSense and AVUT where video durations surpass 1 minute. 
The advantage over direct inference becomes particularly pronounced for videos longer than five minutes, reaching 9.68 points on AVUT and 16.88 points on WorldSense. However, the AVUT $>5$-minute bucket contains only 31 questions, while WorldSense contains 379 questions in the same bucket, providing more reliable evidence for the long-video advantage of agentic inference. In this long-video setting, our compact workflow outperforms the full OmniAgent on AVUT and remains within 0.79 points on WorldSense. Meanwhile, the replicated OmniAgent also obtains larger gains than direct inference with OmniLLM at longer videos. These results further confirm that agentic workflow is more valuable for long-form audio-visual understanding, motivating the following ablations on how to retain its benefits with fewer tools and a more efficient execution trajectory.

\subsection{Essential Tool-Group Selection}
\label{ssec: Tool-Group selection}
To address our first research question—whether the broad specialized toolset can be pruned without sacrificing accuracy, we examines which perceptual capabilities are necessary, shown in table~\ref{tab:stage1_tool_selection}. Removing both the audio-event and ASR groups yields the highest mean accuracy among the ablations (78.93\%), pruning the available tool set to C+Q+M. In contrast, removing question-directed QA substantially reduces tool calls and estimated cost, but accuracy drops sharply to 62.13\%. Removing global audio captioning also lowers accuracy to 75.47\%, demonstrating the value of broad contextual evidence. The minimal C+Q configuration reaches only 74.13\%, suggesting that video metadata, which primarily provides video duration and frame rates, remains useful for temporal grounding.
Overall, these results identify captioning and QA, as two representatives of passive and active evidence acquisition tools respectively,
together with the video metadata tool as providing complementary capabilities, and motivate C+Q+M as the starting configuration for the subsequent modality-consolidation study.

\subsection{Modality Consolidation}
\label{ssec: modality Consolidation}
Table~\ref{tab:stage2_stage3_ablation} traces the transition from specialized tools to our compact omni-tool agent; Systems~0--6 cover modality consolidation.

Systems~0--1 establish a common DSPy implementation. We adopted DSPy for more modularized implementation which allows future automatic tool selections. Because the port also changes serialization and prompt rendering, its difference
from the LangChain result in subsection~\ref{ssec: Tool-Group selection} is treated as a migration gap. Deterministic
Gemini sampling has little effect. System~2 adopts the compact workflow, reducing
calls by 29.0\% and cost by 44.9\%, although losing 3.73 accuracy points.

Systems~3--4 address this bottleneck by replacing audio-only captioning with the
Omni Captioning Tool and then adding the caption-first instruction. Accuracy
recovers to 77.87\% while calls fall to 2.397, indicating that providing captions including evidence from the whole video in an earlier step reduces repeated acquisition.

System~5 consolidates QA roles, further reducing calls and cost but lowering accuracy to 74.40\%. System~6 hides planner-visible metadata and reaches 78.40\% with 1.941 calls, showing that after simply including timestamps in the caption, the video metadata is no longer essential.
From Systems~0 to~6, accuracy rises by 2.93 points, while calls, output tokens, cost, and latency fall by 55.0\%, 58.2\%, 63.3\%, and 66.9\%, respectively. Overall, omni-modal tools and an adapted workflow progressively reduce tool use.

\begin{figure*}[t]
  \centering
  \includegraphics[width=0.9\textwidth]{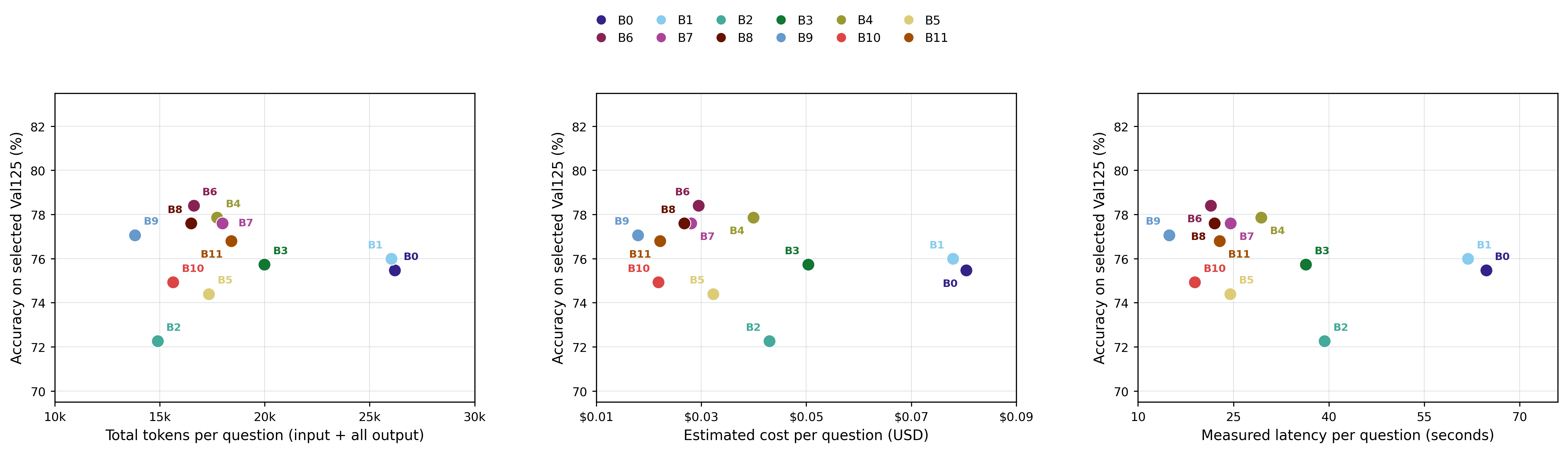}
  \caption{
    Accuracy-efficiency trade-offs across Systems~0--11 on the selected
    Daily-Omni validation subset. The corresponding latency values are
    presented in Appendix~\ref{app:ablation_latency}, to save space for Table \ref{tab:stage2_stage3_ablation}.
  }
  \label{fig:ablation_accuracy_efficiency}
\end{figure*}

\vspace{-0.2cm}
\subsection{Compacting the Omni-Tool Workflow}
\vspace{-0.1cm}

We attempted to further compact the workflow, based on the observation that there is only small partition of long execution question samples, and relatively rare calling of the Omni-modal clip QA tool.
From System~6, we found that although it permitted up to 30 calls, the median call count is 2.00 and roughly 95\% of questions required fewer than 4 calls, motivating the six-call cap in System~7 to eliminate a small number of excessively long execution traces.
System~7 reduces the call budget from 30 to six with only a small change in mean accuracy. System~8 removes clip-level QA while retaining the same 77.60\% mean accuracy, suggesting that it provides limited aggregate benefit once the Omni Captioning Tool and the Omni QA Tool are available. System~9 then lowers planner reasoning effort, retaining 77.07\% accuracy with 1.547 calls and 13.809k total tokens. Relative to System~6, it uses 20.3\% fewer calls and 39.3\% less estimated cost for a 1.33-point accuracy trade-off. Removing the caption-first instruction in System~10 degrades both accuracy and efficiency, further supporting the prompt-level policy for structured initial evidence acquisition.
On the other hand, System~6 offers the best accuracy-efficiency balance in Stage~2. Considering System~6 has the clip-level Omni QA tool, we adds this tool back under System~9's medium-reasoning, six-call setting, resulted in System~11. This yields no accuracy gain over System~9 while consuming slightly more resources; we therefore retain System~9 for its lowest cost and competitive accuracy. Figure \ref{fig:ablation_accuracy_efficiency} also shows System 9 lies on the empirical accuracy-efficiency Pareto frontier.

\section{Discussion}
\label{sec: Discussion}

\subsection{Essential Perceptual Capabilities}
\label{ssec: Essential Perceptual Capabilities}

Strong performance depends on complementary passive and active acquisition rather
than a large toolset: captioning supplies broad context, while QA retrieves and
verifies question-specific evidence. Metadata becomes less useful once the
omni-modal caption provides timestamps. Separate ASR and audio-event tools are
also largely redundant because Gemini supports native audio understanding
\citep{comanici2025gemini}, and our caption prompt already requests speech and
salient sounds.

\subsection{Modality Consolidation and Workflow Co-design}
\label{ssec: Modality Consolidation and Workflow Co-design}

Modality consolidation must be co-designed with the workflow: compacting the
\textit{specialized-tool OmniAgent} pipeline creates an information bottleneck, while replacing the instruction given to the planner model with a more compact workflow instruction does not preserve accuracy. Omni-modal captioning and the caption-first policy recover performance by placing richer evidence in the initial context,
showing that tool roles, order, and evidence placement matter jointly.

Because the Daily-Omni ablations cover only 30--60-second videos, we also test
whether removing clip-level QA generalizes to the longer WorldSense videos,
where temporal narrowing may be more useful
\citep{wang2024videoagent,wang2024videotree,liao2025divide}. Adding the clip
tool back reduces overall accuracy and underperforms in three of four duration
bins (Appendix~\ref{app:worldsense_clip_qa_duration}), possibly because recent
OmniLLMs' ability of identifying relevant periods from the full video has become stronger.
\vspace{-0.2cm}
\section{Conclusions}
\label{sec: conclusions}
\vspace{-0.2cm}
We study whether agentic audio-visual question answering requires many specialized perception tools or can be constructed more compactly with less role and modality specialized tools. Our ablations show that performance depends on two complementary capabilities—broad audio-visual context provided by the captioning tool and question-directed inspection supported by the QA tool, rather than tool count. We therefore develop a two-tool agent combining Omni Captioning and Omni QA with a caption-first policy and a six-time tool call budget.

The resulting system achieves 78.86\%, 80.57\%, and 58.39\% accuracy on Daily-Omni, AVUT, and WorldSense, respectively. It matches or exceeds the reproduced specialized-tool OmniAgent on Daily-Omni and AVUT and remains within 0.91 percentage points on WorldSense, while reducing tool calls and estimated inference cost substantially. Regardless that  OmniLLM itself already performs very strong, the compact agent still improves overall accuracy, which is prominent on videos longer than five minutes. Such results confirm that active evidence acquisition organized by agentic system remains valuable even with a strong OmniLLM as the base perception model. Future work can be carried out focusing more on long audio-visual video understanding and automatic tool selections.

\section*{Limitations}
\label{sec:limitations}

Our study has several limitations. First, because of the substantial cost of repeatedly evaluating API-based agentic systems, the controlled ablations are repeated only on the 125-example Daily-Omni validation subset, while the complete benchmark evaluations are conducted with a single run. Repeated ablations reveal non-negligible run-to-run variance in both accuracy and efficiency, particularly for some intermediate configurations. Consequently, small differences between configurations should not be interpreted as statistically conclusive, and repeated full-benchmark evaluation would provide more reliable estimates.

Second, the main experiments use a single text-only planner, o3, and a single shared perception backbone, Gemini-2.5-Flash. Both are closed-source, API-accessible models whose behavior, availability, and pricing may change over time, limiting reproducibility and making it difficult to separate general workflow properties from model-specific behavior. 

Third, the joint design space of clip-level QA availability, tool-call budget, and planner reasoning effort is not explored through a formal greedy or exhaustive search. Instead, these factors are modified sequentially during the Stage 2 and Stage 3 ablations. 
We therefore do not claim that the final configuration is globally optimal. Our primary aim is to maintain competitive accuracy while identifying the lowest-cost workflow among the evaluated configurations, under which criterion the selected two-tool system provides the strongest overall balance. A larger factorial or multi-objective search could more precisely characterize the Pareto frontier between accuracy, and efficiency, on various scenarios (e.g. on various lengths of videos).

Finally, the tool-selection process remains partly heuristic. Functional groups are manually defined according to the roles of the original specialized tools, and subsequent configurations are chosen based on staged empirical observations. Although this procedure makes the capability-removal process interpretable, it may overlook alternative tool combinations or interfaces and depend on the initial tool taxonomy. Future work could formulate tool consolidation as an automated architecture-selection problem.

\section*{Ethical Considerations}

This work uses existing audio-visual question answering benchmarks and does not collect new human-subject data or release new video data. The main risks arise from errors in audio-visual perception and reasoning. Generated captions or question-directed observations may omit relevant evidence, hallucinate events, misinterpret speech, or incorrectly associate audio with visual content. Such errors could cause harm if the framework were applied in high-stakes decision-making, monitoring, or surveillance-like settings. Our experiments are limited to benchmark evaluation, and the system is not intended for high-stakes deployment without additional validation and appropriate human oversight.

The study relies on large API-based models, which incur computational, financial, and environmental costs. A central motivation of this work is to reduce unnecessary inference by pruning redundant tools, consolidating modality-specific functions, and limiting the number of valid tool calls. The resulting compact workflow substantially reduces tool interactions, token consumption, and estimated inference cost relative to the reproduced specialized-tool agent. We report the model configurations, reasoning settings, tool-call budgets, and efficiency measurements to support transparent comparison and reproducibility.

The benchmark datasets used in this study are publicly accessible. The o3 and
Gemini-2.5-Flash model weights are proprietary and are accessed through their
respective APIs. We use all datasets, models, and software artifacts only for
research and benchmark evaluation and follow their corresponding licenses and
terms of use.


\bibliography{custom}

\appendix

\section{WorldSense Clip-Level QA Ablation by Video Duration}
\label{app:worldsense_clip_qa_duration}
WorldSense clip-Level QA ablation by video duration is shown in table \ref{tab:worldsense_clip_qa_duration}.
\begin{table*}[t]
  \centering
  \small
  \setlength{\tabcolsep}{6pt}
  \renewcommand{\arraystretch}{1.04}
  \begin{tabular*}{\textwidth}{@{\extracolsep{\fill}}lrrrrr@{}}
    \toprule
    Method & Overall & $\leq 1$ min & 1--3 min & 3--5 min & $>5$ min \\
    \midrule
    OmniAgent (reproduced)
      & \textbf{59.30} & 57.13 & \textbf{60.82} & 54.02 & \textbf{64.64} \\
    ours (w/o Omni clip tool)
      & 58.39 & 55.43 & 59.52 & \textbf{56.90} & 63.85 \\
    extended (+ clip-level Omni QA)
      & 57.79 & \textbf{57.22} & 58.51 & 53.16 & 60.95 \\
    \bottomrule
  \end{tabular*}
  \caption{WorldSense accuracy (\%) by video duration for the reproduced
  specialized-tool OmniAgent, our two-tool system, and the
  same system with the clip-level Omni QA tool added back.
  The duration buckets contain 1,059, 1,386, 348, and 379 questions,
  respectively.}
  \label{tab:worldsense_clip_qa_duration}
\end{table*}

\section{Implementation Details}
\label{app:implementation_details}

\subsection{Model and Inference Configuration}

Table~\ref{tab:model_inference_configuration} reports the inference settings
of our final two-tool system. Although the planner configuration records a
temperature of zero, the o3 adapter omits temperature from the actual API
request because this parameter is not exposed for the model. The first
\texttt{Omni global caption} observation is read from a cache generated with the same Gemini model, sampling parameters, and caption prompt reported below;
subsequent tool calls are executed live.
For fair per-question efficiency accounting, the reported token and cost metrics
attribute one complete caption-generation call to every question, including
questions that share a source video. The reported efficiency values are therefore
caption-inclusive and do not discount caption generation merely because the
caption was cached. Ablation latency was recorded during the live rollout and
includes the captioning step; no post-hoc latency reconstruction is used.

\begin{table*}[t]
  \centering
  \small
  \setlength{\tabcolsep}{5pt}
  \renewcommand{\arraystretch}{1.08}
  \begin{tabular*}{\textwidth}{@{\extracolsep{\fill}}llrrrrrr@{}}
    \toprule
    Component & Model & Reasoning & Temp. & Top-$p$ & Top-$k$
      & Max output & Seed \\
    \midrule
    Planner & o3 & medium & not sent & -- & -- & 32,768 & 1234 \\
    Captioning and QA tools & Gemini-2.5-Flash & provider default
      & 0.0 & 1.0 & 0 & 4,096 & 1234 \\
    \bottomrule
  \end{tabular*}
  \caption{Model and sampling parameters for the proposed system. Maximum
  output lengths are measured in tokens.}
  \label{tab:model_inference_configuration}
\end{table*}

The planner is allowed at most six valid tool calls. The required first
caption counts toward this budget, whereas the final-answer planner round does
not. Gemini requests use a 180-second timeout, at most four total attempts,
and a 30-second delay between attempts. The planner and both perception tools
use the same settings across Daily-Omni, AVUT, and WorldSense.

Across the reported experiments and associated development runs, API expenditure
was approximately \$500 for o3 and \$100 for Gemini-2.5-Flash. Both models are
closed-source, and their parameter counts and provider-side inference hardware
are not disclosed. Local computation was used to orchestrate API requests and
process benchmark media rather than to host the model weights.

Table~\ref{tab:software_versions} reports the principal software packages used
to implement the proposed pipeline and reproduce OmniAgent.

\begin{table}[t]
  \centering
  \small
  \setlength{\tabcolsep}{5pt}
  \renewcommand{\arraystretch}{1.08}
  \begin{tabular}{lll}
    \toprule
    Package & Version & Role \\
    \midrule
    DSPy & 3.2.1 & Proposed pipeline \\
    LiteLLM & 1.82.6 & Model/API routing \\
    LangChain & 1.1.0 & OmniAgent reproduction \\
    \bottomrule
  \end{tabular}
  \caption{Principal software versions used in the experiments.}
  \label{tab:software_versions}
\end{table}

\section{Ablation Latency}
\label{app:ablation_latency}

Table~\ref{tab:ablation_latency} reports the per-question retry-adjusted
latency for Systems~0--11.

\begin{table}[t]
  \centering
  \small
  \setlength{\tabcolsep}{6pt}
  \renewcommand{\arraystretch}{1.05}
  \begin{tabular}{rrrr}
    \toprule
    System & Latency (s) $\downarrow$ & System & Latency (s) $\downarrow$ \\
    \midrule
    0 & $64.73 \pm 8.64$ & 6  & $21.43 \pm 2.95$ \\
    1 & $61.84 \pm 7.02$ & 7  & $24.54 \pm 2.26$ \\
    2 & $39.33 \pm 9.55$ & 8  & $22.03 \pm 3.13$ \\
    3 & $36.37 \pm 9.43$ & 9  & $14.90 \pm 2.40$ \\
    4 & $29.35 \pm 6.65$ & 10 & $18.93 \pm 4.98$ \\
    5 & $24.50 \pm 6.45$ & 11 & $22.81 \pm 1.56$ \\
    \bottomrule
  \end{tabular}
  \caption{Latency (excluding error handling retry) per question for the controlled ablations.
  Values are mean $\pm$ sample standard deviation over three runs.}
  \label{tab:ablation_latency}
\end{table}

\subsection{Prompts for the Proposed System}

Tables~\ref{tab:planner_prompt} and~\ref{tab:tool_prompts} present the key
instructions used by the final system, lightly formatted for readability. \textit{ask\_perception} is the omni-modal QA tool on whole video, and \textit{ask\_caption} is the omni-modal captioning tool on whole video.

\begin{table*}[p]
\centering
\small
\renewcommand{\arraystretch}{1.15}
\begin{tabular}{p{0.95\linewidth}}
\toprule

\textbf{Role and Accessible Evidence:}

You are the planning/reasoning model in an audio-visual question answering
ReAct system. You CANNOT watch the video directly. You can only reason over
(1) the user question and options, (2) observations returned by the captioner
tool, and (3) observations returned by the perceptual tool.

\\
\midrule

\textbf{Tool Interfaces:}

There are two different tools. \texttt{ask\_caption} calls the captioner model;
use it only to obtain a factual timestamped audio-visual caption of the whole
video. \texttt{ask\_perception} calls the perceptual model; use it only after
reading the caption, to ask any audio-visual perception question that may help
answer the user question.

\\
\midrule

\textbf{Required Workflow:}

(1) The first tool call must be \texttt{ask\_caption} so that you can observe
what happens in the video.

(2) From the second tool call onward, freely choose \texttt{ask\_caption} or
\texttt{ask\_perception} based on the evidence you still need. You may call
either tool multiple times.

(3) Return the final answer as soon as the available evidence is sufficient.

\\
\midrule

\textbf{Tool-Call Freedom after the First Caption:}

\texttt{ask\_perception} may ask any audio, visual, or audio-visual perception
question. The perceptual question may quote, reuse, or directly ask the
original user question and options, and does not have to be narrowly targeted
or limited to missing or ambiguous evidence. \texttt{ask\_caption} may be
called at any later step and is not limited to one call.

\\
\midrule

\textbf{Budget and Constraints:}

Each decision includes \texttt{turn\_index} and \texttt{max\_turns}.
\texttt{max\_turns} is the total tool-call budget, including the required first
\texttt{ask\_caption} call. You may make at most \texttt{max\_turns} tool calls.
If the budget is exhausted, you will be required to return a final answer
using the evidence already collected. Do not call \texttt{ask\_perception}
before the first \texttt{ask\_caption} observation. For
\texttt{ask\_caption}, request the default factual whole-video caption by using
empty arguments. Do not output a final answer before receiving the first
caption observation. Do not output Markdown fences.

\\
\midrule

\textbf{Action Schema:}

Return exactly one JSON object. For a caption observation, return
\texttt{\{"action": "tool", "tool\_name": "ask\_caption", "arguments": \{\}\}}.
For a perceptual observation, return
\texttt{\{"action": "tool", "tool\_name": "ask\_perception",}
\texttt{"arguments": \{"perceptual\_question": "..."\}\}}.
When the evidence is sufficient, return
\texttt{\{"action": "final", "answer": "<single option label>",}
\texttt{"reasoning\_summary": "<brief explanation>"\}}. Return JSON only.

\\
\midrule

\textbf{Budget-Exhaustion Instruction:}

Max turns reached. Do not call any tool. Return exactly one final JSON object
now:
\texttt{\{"action": "final", "answer": "<single option label>",}
\texttt{"reasoning\_summary": "<brief explanation>"\}}.

\\
\bottomrule
\end{tabular}
\caption{Planner workflow and action-schema prompt used by the proposed
system.}
\label{tab:planner_prompt}
\end{table*}

\begin{table*}[p]
\centering
\small
\renewcommand{\arraystretch}{1.15}
\begin{tabular}{p{0.95\linewidth}}
\toprule

\textbf{Omni Captioning Tool:}

You are an audio-visual captioning system. Given a video, produce a detailed
timestamped shot list describing all salient events.

Requirements: Be 100\% factual; describe all salient events and changes that
is visible or audible. Include: Visual elements (scenes, objects, actions);
Audio elements (sounds, music); Speech content (who speaks and what is said,
if clear). Provide timestamps for each event.

\\
\midrule

\textbf{Omni QA Tool:}

Answer the question directly using only observable video/audio evidence. Do
not guess beyond evidence.

Question: \texttt{\{perceptual\_question\}}

Return JSON with keys \texttt{answer} and \texttt{evidence}.

\\
\bottomrule
\end{tabular}
\caption{Prompts for the two tools in the proposed system. Both tools use
Gemini-2.5-Flash with the same audio-visual input; their roles differ through
these instructions.}
\label{tab:tool_prompts}
\end{table*}

\section{Daily-Omni Split Statistics}
\label{app:daily_omni_statistics}

The 125-example validation subset used for all controlled ablations was
stratified by the six Daily-Omni task types. As shown in
Table~\ref{tab:daily_omni_task_distribution}, its task distribution closely
matches that of the complete benchmark.

\begin{table*}[t]
  \centering
  \small
  \setlength{\tabcolsep}{5pt}
  \renewcommand{\arraystretch}{1.05}
  \begin{tabular*}{\textwidth}{@{\extracolsep{\fill}}lrr@{}}
    \toprule
    Task type & Full benchmark & Val125 \\
    \midrule
    AV Event Alignment & 238 (19.88\%) & 24 (19.20\%) \\
    Comparative & 131 (10.94\%) & 14 (11.20\%) \\
    Context Understanding & 193 (16.12\%) & 20 (16.00\%) \\
    Event Sequence & 306 (25.56\%) & 33 (26.40\%) \\
    Inference & 154 (12.87\%) & 16 (12.80\%) \\
    Reasoning & 175 (14.62\%) & 18 (14.40\%) \\
    \midrule
    Total & 1,197 (100\%) & 125 (100\%) \\
    \bottomrule
  \end{tabular*}
  \caption{Question counts and proportions by task type in the complete
  Daily-Omni benchmark and the stratified validation subset used for the
  controlled ablations.}
  \label{tab:daily_omni_task_distribution}
\end{table*}

\section{Benchmark and Artifact Documentation}
\label{app:benchmark_documentation}

Table~\ref{tab:benchmark_documentation} summarizes the composition, provenance,
and documentation relevant to the three evaluation benchmarks. We use only the
released evaluation data and do not collect or redistribute any new videos. We
did not conduct an additional independent audit for personally identifying or
offensive content, and instead rely on the curation and release procedures
reported by the benchmark creators.

\begin{table*}[t]
  \centering
  \small
  \setlength{\tabcolsep}{5pt}
  \renewcommand{\arraystretch}{1.08}
  \begin{tabular}{p{0.13\textwidth}p{0.35\textwidth}p{0.43\textwidth}}
    \toprule
    Benchmark & Composition used in this work & Provenance and documentation \\
    \midrule
    Daily-Omni
      & 684 videos and 1,197 English multiple-choice questions across six task
        types; clips span 30 or 60 seconds.
      & Videos are drawn from AudioSet, Video-MME, and FineVideo. The dataset
        paper describes human review of generated questions but does not report
        a separate PII or offensive-content audit. \\
    \midrule
    AVUT
      & The AV-Human partition: 698 videos and 1,734 expert-annotated English
        questions across 18 audio-centric domains.
      & Videos were collected from publicly available YouTube content. The
        release provides links rather than hosting or redistributing the video
        files and directs users to comply with YouTube's terms of service. \\
    \midrule
    WorldSense
      & 1,662 audio-visual videos and 3,172 multiple-choice questions spanning
        8 domains, 67 subcategories, and 26 tasks, annotated by 80 professional
        annotators.
      & The dataset paper reports collection from public sources, removal of
        personally identifying information, and a second manual quality check
        for ethical violations. It is intended for academic evaluation of
        real-world omni-modal understanding. \\
    \bottomrule
  \end{tabular}
  \caption{Documentation of the benchmark artifacts used in this work. Counts
  and curation statements follow the corresponding dataset papers
  \citep{zhou2025daily,yang2025audio,hong2025worldsense}.}
  \label{tab:benchmark_documentation}
\end{table*}

\section{Video-Disjoint Daily-Omni Held-Out Evaluation}
\label{app:daily_omni_heldout}

Because the heuristic toolset selection was conducted on Val125, we also
evaluate the three Daily-Omni systems from Table~\ref{tab:main_cross_benchmark}
on a video-disjoint held-out partition. We identify the 119 unique videos
represented in Val125 and exclude \emph{all} questions associated with those
videos from the complete benchmark, rather than excluding only the 125
validation question IDs. This leaves 956 questions from the other 565 videos;
the exclusion removes an additional 116 questions that share videos with
Val125.

\begin{table*}[t]
  \centering
  \small
  \setlength{\tabcolsep}{7pt}
  \renewcommand{\arraystretch}{1.05}
  \begin{tabular*}{\textwidth}{@{\extracolsep{\fill}}lrr@{}}
    \toprule
    Method & Whole benchmark & Video-disjoint held-out \\
    \midrule
    Gemini-2.5-Flash & 77.03 (922/1,197) & 77.09 (737/956) \\
    OmniAgent (reproduced) & 77.53 (928/1,197) & 77.30 (739/956) \\
    Ours & \textbf{78.86 (944/1,197)} & \textbf{79.08 (756/956)} \\
    \bottomrule
  \end{tabular*}
  \caption{Daily-Omni accuracy (\%) on the complete benchmark and after
  excluding every video represented in the Val125 development subset.}
  \label{tab:daily_omni_video_disjoint}
\end{table*}

Our compact system retains the highest accuracy on the video-disjoint
partition, and its accuracy does not decrease relative to the whole-benchmark
result. This suggests that the heuristic toolset selection did not overfit to
the videos used in the controlled ablations.

\end{document}